\documentclass[letterpaper]{article}
\usepackage{spconf,amsmath,amssymb,graphicx}
\usepackage{booktabs,cite,xspace,tikz,flushend}
\usepackage[stretch=25]{microtype}
\usepackage[hidelinks]{hyperref}
\newcommand{\allot}{\textsc{ALLOT}\xspace}
\newcommand{\ext}{\mathrm{ext}}
\newcommand{\both}{\mathrm{both}}
\newcommand{\para}{\mathrm{para}}
\newcommand{\acc}{\operatorname{acc}}

\title{ALLOT: Budgeted Hybrid-Memory Routing for Knowledge Updates in LLMs}
\name{Shanfeng Huang, Zhou Fang, Song Xiao, Hai Du}
\address{Baidu Inc.\\\{huangshanfeng, fangzhou07, xiaosong, duhai02\}@baidu.com}
\hypersetup{pdftitle={ALLOT: Budgeted Hybrid-Memory Routing for Knowledge Updates in LLMs},pdfauthor={Shanfeng Huang, Zhou Fang, Song Xiao, Hai Du}}
\begin{document}
\raggedbottom
\maketitle
\begin{abstract}
For large language models (LLMs), parametric adaptation is costly when retrieval already suffices. We introduce ALLOT, a hybrid-memory routing framework that separates learned write priority from a hard parametric budget. A memory-aware router combines frozen text representations, retrieval confidence, and relation metadata; a single ranking supports multiple write budgets while preserving all facts in external memory. On CounterFact with Qwen3-4B, ALLOT reaches 0.760 accuracy at a 20\% parametric-write budget and recovers 78.4\% of the budget-matched oracle gain, with 80\% fewer parametric writes than dual-writing every fact. At this budget, jointly adding retrieval and relation features to text improves normalized oracle gain by 6.2 percentage points. Complementary Qwen3-0.6B shared-store results achieve dual-write-level accuracy with 6--14.5\% parametric writes, and cross-benchmark transfer retains approximately 88\% of in-domain gain. These results support allocating adaptation capacity according to its incremental value rather than treating every factual update as an equally valuable training target.
\end{abstract}
\section{Introduction}
\label{sec:intro}
Large language models (LLMs) can access new facts through retrieval or internalize them through parameter updates. External writes are inexpensive; training and retaining a LoRA adapter adds compute and storage. Writing every fact to both memories therefore spends resources on facts already answered by retrieval. With limited resources, the allocation question is \emph{which facts deserve an additional parametric write}.

Retrieval-augmented models\,\cite{lewis2020rag,guu2020realm,izacard2023atlas} exploit external knowledge, while ROME, MEMIT, and SERAC\,\cite{meng2022rome,meng2022memit,mitchell2022serac} update factual behavior. LoRA\,\cite{hu2022lora} reduces adaptation cost, and AdaLoRA\,\cite{zhang2023adalora} allocates adaptation rank. WISE\,\cite{wang2024wise} routes between main and side parametric memories for lifelong editing; SPRInG\,\cite{kim2026spring} combines drift-driven adaptation with retrieval-interpolated personalization. Parametric RAG turns retrieved documents into adapters, one per document, and dynamic variants amortize that cost through a parameter translator\,\cite{su2025prag,tan2025dyprag}. Such designs decide how memories are formed, read, or combined; \allot addresses an orthogonal question: under a global adaptation budget, which updates should be written parametrically in the first place. This is related to learning to defer\,\cite{madras2018learningtodefer}, with storage actions replacing predictor choices.

The challenge is not simply to detect difficult facts. Poor retrieval does not itself imply that adaptation helps. A useful router must prioritize the \emph{incremental value} of the second memory. A fixed action classifier cannot enforce a workload-level write budget. Ranking separates this learned preference from the available write capacity.

\begin{figure}[t]
\centering
\begin{tikzpicture}[x=1cm,y=1cm,>=stealth,font=\small,
 box/.style={draw,rounded corners=2pt,align=center,minimum height=.7cm,inner sep=5pt}]
\node[box,text width=3cm] (features) at (1.5,0)
 {Text embedding\\Retrieval confidence\\Relation metadata};
\node[box,text width=2.5cm] (router) at (5.2,0)
 {\allot router\\Score $s_i$};
\node[box,text width=2.5cm] (budget) at (5.2,-1.3)
 {Rank items\\Select top $B$};
\node[box,text width=3cm] (external) at (1.5,-2.6)
 {External index\\All $N$ items};
\node[box,text width=2.5cm] (adapter) at (5.2,-2.6)
 {Adapter bank\\Top $B$ items};
\draw[->] (features) -- (router);
\draw[->] (router) -- (budget);
\draw[->] (budget) -- (adapter);
\draw[->] (features) -- node[right,align=left]{write every\\fact} (external);
\end{tikzpicture}
\caption{\allot allocation retains every fact in external memory and reserves parametric writes for the highest router scores. Oracle probe outcomes supervise training only; changing the budget changes the selection cutoff.}
\label{fig:routing}
\end{figure}
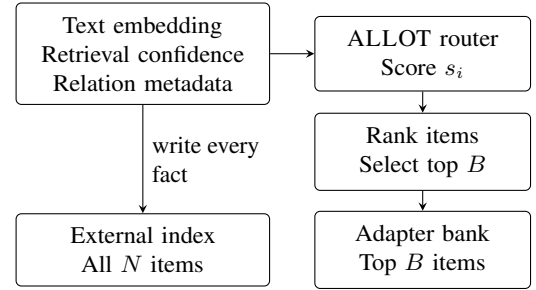

\allot implements this separation through a memory-aware priority score and a deterministic budget cutoff. Every fact receives an external write; only the top-ranked facts receive adapters. Text embeddings describe the fact, retrieval statistics characterize its memory neighborhood, and relation metadata provides its semantic category. The budget controls the number of facts adapted, independently of retrieval-store size and adapter rank. One ranking serves multiple operating points without retraining or selecting a new probability threshold.

Our contributions are threefold. \textbf{(1) Budget-controlled hybrid memory:} we formulate selective parametric writing as constrained allocation and give a reusable-ranking procedure with exact write-count control and nested selections across budgets. \textbf{(2) Memory-aware prioritization:} we combine action-outcome supervision with text, retrieval, and relation signals, and characterize label stability and allocation error. \textbf{(3) An empirical allocation frontier:} CounterFact/4B results recover 78.4\% of oracle gain at 20\% budget; feature ablations, shared-store results, and cross-benchmark transfer reveal when selective writing is effective.

\section{Budgeted Hybrid-Memory Routing}
\label{sec:method}

\subsection{Allocation objective}
For each factual item $x_i$, three question--answer probes evaluate four actions: no write (drop), retrieval only (ext), adapter only (para), and both memories (both). Let $\acc_i(a)$ be the fraction of generated answers containing the gold entity, ignoring case. Action outcomes are recorded per item and action from the same initial memory state.

The hard-budget policy retains all $N$ items externally and permits $B=\lfloor bN\rfloor$ parametric writes. With $z_i\in\{0,1\}$ and $\Delta_i=\acc_i(\both)-\acc_i(\ext)$, it seeks
\begin{equation}
\max_{z}\frac{1}{N}\sum_{i=1}^{N}[\acc_i(\ext)+z_i\Delta_i],
\quad \sum_{i=1}^{N}z_i\le B.
\label{eq:objective}
\end{equation}
For fixed per-item outcomes, the capped oracle selects up to $B$ largest positive gains; an exact-$B$ oracle instead selects the $B$ largest gains even if some are negative. We use the capped oracle as an upper reference for the exact-$B$ router. This fixed-outcome objective makes allocation quality the only free variable.

\subsection{Supervision and feature representation}
Supervision labels maximize accuracy minus write costs:
\begin{equation}
a_i^*=\arg\max_a[\acc_i(a)-\lambda_p r_p(a)-\lambda_e r_e(a)],
\label{eq:oracle}
\end{equation}
where $r_p,r_e$ indicate parametric and external writes and $\lambda_p=\lambda_e=0.1$. Probe outcomes provide supervision, not input features. The four labels distinguish which memories support each fact.

\noindent\textbf{Label stability.} With three binary probes, unequal accuracies differ by at least $1/3$. For equal penalties $0<\lambda_p=\lambda_e=\lambda<1/6$, the largest cost difference, $2\lambda$, is smaller than this gap. Accuracy therefore determines all unequal-accuracy comparisons; fewer writes break equal-accuracy ties. With fixed remaining tie breaking, labels are unchanged throughout this interval, which contains $\lambda=0.1$. The penalty shapes supervision; the cutoff, not $\lambda$, enforces the deployment budget.

Inputs concatenate three $D$-dimensional slots: mean-pooled frozen text embeddings; cosine-similarity statistics over the top-5 retrieved neighbors, zero-padded; and zero-padded relation one-hot encodings. Ablations zero inactive slots. An MLP with two ReLU hidden layers ($3D\!\rightarrow\!512\!\rightarrow\!128\!\rightarrow\!4$) minimizes class-weighted cross-entropy. Weights are inversely proportional to class frequency, countering the 53\% ext-label majority on CounterFact/4B. For training set $\mathcal{T}$, the loss is
\begin{equation}
\mathcal{L}(\theta)=-\frac{1}{|\mathcal{T}|}\sum_{i\in\mathcal{T}}
w_{a_i^*}\log p_\theta(a_i^*\mid x_i).
\label{eq:training}
\end{equation}
The encoder remains frozen. Zeroed slots keep the architecture fixed across feature modes. For $D=2560$, the routing head has approximately 4.0 million trainable parameters, excluding the encoder and memory adapters.

\subsection{Budgeted routing algorithm}
The score aggregates both parametric-write labels:
\begin{equation}
s_i=p_\theta(a_i^*=\para\mid x_i)+p_\theta(a_i^*=\both\mid x_i).
\label{eq:score}
\end{equation}
For the ext/both decision, the utility difference in Eq.~\eqref{eq:oracle} is $\Delta_i-\lambda_p$: external-write cost cancels. By that comparison an item carries the both label only when $\Delta_i>\lambda_p$, so this probability mass tracks strictly positive marginal gain, while the para mass marks facts the adapter alone already answers. Their sum supplies the ranking, and measured marginal gains determine allocation quality across budgets.

\noindent\textbf{Allocation procedure.}
Train on the four-action labels; score each candidate once; sort by decreasing $s_i$; write every item externally and the top $B$ items parametrically. Ties use a fixed item order, giving nested selections as $B$ changes. The deployed policy therefore uses exactly $B$ parametric writes at every operating point. Let $\pi$ be the resulting permutation of item indices. The deployed storage action is
\begin{equation}
\widehat a_{\pi(k)}(B)=
\begin{cases}
\both, & k\le B,\\
\ext, & k>B.
\end{cases}
\label{eq:allocation}
\end{equation}
Sorting costs $O(N\log N)$; changing the budget changes only the cutoff for a fixed candidate pool and feature context. The extremes are retrieval-only at $B=0$ and dual-write at $B=N$.

\subsection{Why ranking quality matters}
Let $S_B$ be the router's selection and $O_B$ the oracle's. Their accuracies under Eq.~\eqref{eq:objective} satisfy
\begin{equation}
A_{O_B}-A_{S_B}=\frac{1}{N}\left(
\sum_{i\in O_B\setminus S_B}\Delta_i-
\sum_{i\in S_B\setminus O_B}\Delta_i\right).
\label{eq:gap_decomp}
\end{equation}
The first term measures missed oracle-selected gains, and the second credits gains from alternative selections. Allocation error thus depends on which items cross the budget cutoff, not four-way classification accuracy alone. Confusing para with both leaves the aggregated score unchanged, while promoting a low-gain item can displace a higher-gain write. This decomposition links the reusable ranking to the budget--accuracy frontier: an additional write changes accuracy by its realized $\Delta_i/N$, which can be positive or negative.

\section{Experimental Setup}
\label{sec:setup}

\noindent\textbf{Data and memories.}
We use 500 factual updates each from zsRE\,\cite{levy2017zsre} and CounterFact\,\cite{meng2022rome}, with three paraphrase probes per item, and Qwen3-0.6B/4B\,\cite{yang2025qwen3}. External memory uses a flat FAISS index\,\cite{johnson2019faiss} with top-5 cosine retrieval over mean-pooled last-layer representations. Rank-16 LoRA adapters target query/value projections; read-time selection uses a flat cosine-similarity scan over fact embeddings. The 0.6B setup packs up to eight facts per adapter and trains for three epochs; 4B uses individual adapters and 30 steps, both at learning rate $2\times10^{-4}$.

\noindent\textbf{Budget-allocation protocol.}
The main 4B study uses the per-item budget protocol over three 80/20 partitions (42, 123, 456). Comparators are random allocation (50 seeds), balanced logistic regression, GBT (100 estimators, depth 5), and Ridge regression on $\Delta_i$ (uplift), each fit on the same oracle outcomes and applied through the same top-$B$ rule. Feature modes A--D hold the routing architecture fixed to isolate feature contributions. We summarize each configuration by its per-partition means and paired differences across the three splits.

\noindent\textbf{Metrics and complementary protocols.}
Accuracy measures factual recall using the substring metric in Section~\ref{sec:method}. For partition-mean accuracies at a common budget, normalized oracle gain is
\begin{equation}
\text{OG}(b)=\frac{A_{\text{router},b}-A_{\text{random},b}}
{A_{\text{oracle},b}-A_{\text{random},b}}\times\text{100}.
\label{eq:og}
\end{equation}
OG uses unrounded means; a zero denominator makes it undefined. All reported budgets have positive denominators. Complementary 0.6B results use a shared store, evaluating probes after each write; transfer trains on zsRE and evaluates on CounterFact without target-benchmark labels. The 4B sequential protocol tests budget pacing over different arrival orders.

\section{Results}
\label{sec:results}

\subsection{Accuracy under a hard write budget}
\begin{figure*}[t]
\centering
\includegraphics[width=\textwidth]{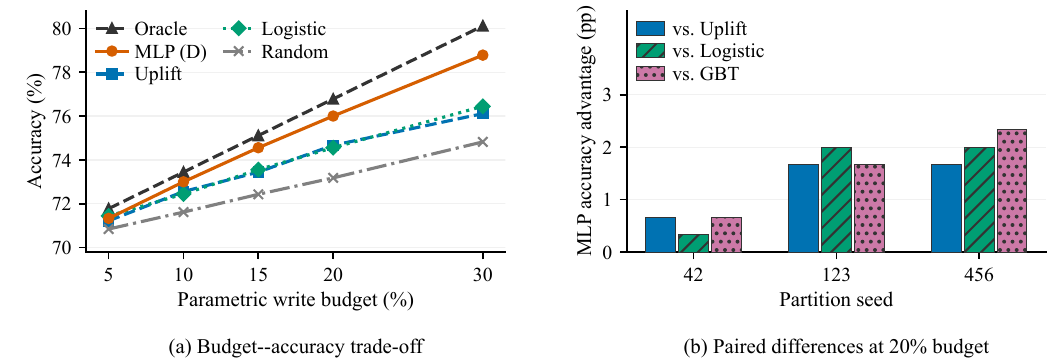}
\caption{CounterFact/Qwen3-4B budget allocation results. (a) Mean accuracy over three partitions; the same \allot scores support every budget. (b) \allot minus baseline accuracy within each partition at 20\% budget. Positive differences occur in all three partitions. Protocol and feature configurations are specified in Section~\ref{sec:setup}.}
\label{fig:budget_analysis}
\end{figure*}

\begin{table}[t]
\caption{CounterFact/Qwen3-4B per-item accuracy (means of three partitions). The oracle uses the same write budget. Protocol and comparator details are specified in Section~\ref{sec:setup}.}
\label{tab:budget}
\centering\small
\setlength{\tabcolsep}{3pt}
\begin{tabular}{@{}lccccc@{}}
\toprule
Method & 5\% & 10\% & 15\% & 20\% & 30\% \\
\midrule
Oracle & 0.718 & 0.734 & 0.751 & 0.768 & 0.801 \\
\allot (D) & 0.713 & \textbf{0.730} & \textbf{0.746} & \textbf{0.760} & \textbf{0.788} \\
Uplift & 0.712 & 0.726 & 0.734 & 0.747 & 0.761 \\
Logistic & \textbf{0.714} & 0.724 & 0.736 & 0.746 & 0.764 \\
GBT & 0.713 & 0.723 & 0.733 & 0.744 & 0.759 \\
Random & 0.708 & 0.716 & 0.724 & 0.732 & 0.748 \\
\bottomrule
\end{tabular}
\end{table}

Table~\ref{tab:budget} and Fig.~\ref{fig:budget_analysis}(a) show the budget--accuracy frontier. At 20\% budget, \allot reaches 0.760 accuracy, versus 0.732 for random, 0.747 for uplift, and 0.768 for the budget-matched oracle. From unrounded means, \allot recovers 78.4\% of oracle gain, compared with 41.3\% for uplift and 35.1\% for GBT. The 1.33 pp accuracy advantage over uplift corresponds to 37.1 pp more normalized oracle gain. These are gains over random at the same write budget.

\allot leads the listed non-oracle configurations at 10--30\% budgets; logistic regression is strongest at 5\%. At 20\%, \allot exceeds uplift by 0.67--1.67 percentage points (pp) and GBT by 0.67--2.33 pp across the three partitions (Fig.~\ref{fig:budget_analysis}(b)). The advantage has the same sign across all three partitions.

\noindent\textbf{Equal-quality write savings.}
\allot at 15\% and logistic regression at 20\% both reach 0.74556 before rounding: \allot uses 25\% fewer parametric writes for the same recorded accuracy. At 20\%, it is within 0.11 pp of uplift at 30\%, with one-third fewer writes. These comparisons quantify the resource value of the learned ordering.

Accuracy and OG answer different questions. From 20\% to 30\% budget, \allot accuracy rises from 0.760 to 0.788, while OG falls from 78.4\% to 74.8\% because the oracle's improvement over random grows faster.

\subsection{Joint value of routing features}
\begin{table}[t]
\caption{Feature ablation on CounterFact/Qwen3-4B. Values are normalized oracle gain (\%) above random at each budget; bold indicates the best feature mode.}
\label{tab:features}
\centering\small
\setlength{\tabcolsep}{3pt}
\begin{tabular}{@{}lccccc@{}}
\toprule
Features & 5\% & 10\% & 15\% & 20\% & 30\% \\
\midrule
A: text & 52.2 & 57.4 & 62.7 & 72.2 & 68.5 \\
B: text + ret. & \textbf{64.1} & 63.5 & 66.8 & 69.1 & 68.5 \\
C: text + rel. & 40.3 & 51.3 & 62.7 & 62.9 & 64.3 \\
D: all & 52.2 & \textbf{75.7} & \textbf{79.3} & \textbf{78.4} & \textbf{74.8} \\
Ret. heuristic & 28.4 & 26.9 & 12.9 & 4.3 & 1.3 \\
\bottomrule
\end{tabular}
\end{table}

At 20\% budget, the routing head already extracts a strong ranking from text alone: mode A reaches 72.2\% OG (Table~\ref{tab:features}), above the uplift and classification rankers in Table~\ref{tab:budget}, and the joint feature set adds a further 6.2 pp (mode D). Adding retrieval confidence alone (B: 69.1\%) or relation metadata alone (C: 62.9\%) lowers the text-only value, whereas their joint inclusion (D: 78.4\%) is strongest. A retrieval-confidence heuristic reaches only 4.3\%. Joint text, memory context, and relation signals outperform direct confidence ranking at this operating point.

Mode D also exceeds text alone by 18.3 pp OG at 10\% budget and 16.6 pp at 15\%. Text plus retrieval is best at 5\%, whereas the full combination is strongest at 10--30\%. Feature utility therefore depends on the selected budget cutoff.

\subsection{Shared-store efficiency and transfer}
\begin{table}[t]
\caption{Qwen3-0.6B shared-store results, 500 items per dataset. Para. is the fraction of parametric writes; MB values are archived storage estimates. \allot accuracy is a three-seed mean.}
\label{tab:efficiency}
\centering\small
\setlength{\tabcolsep}{3pt}
\begin{tabular}{@{}llccc@{}}
\toprule
Data & Policy & Acc. & Para. & MB (est.) \\
\midrule
zsRE & Retrieval & .624 & 0\% & 2 \\
& Both always & .644 & 100\% & 553 \\
& \allot & .646 & 6\% & 37 \\
\midrule
CF & Retrieval & .516 & 0\% & 2 \\
& Both always & .585 & 100\% & 553 \\
& \allot & .585 & 14.5\% & 63 \\
\bottomrule
\end{tabular}
\end{table}

The complementary 0.6B shared-store runs match dual-write accuracy with substantially fewer adaptations (Table~\ref{tab:efficiency}): 0.646 and 0.585, on par with the 0.644 and 0.585 of dual-writing, using only 6\% and 14.5\% parametric writes. Archived storage estimates are 553 MB for dual-writing and 37/63 MB for routing, corresponding to estimated reductions of 14.9$\times$/8.8$\times$.

Cross-benchmark evaluation further reports 0.577 CounterFact accuracy for a router trained on zsRE, versus 0.516 for retrieval and 0.585 for in-domain routing. This retains $(0.577-0.516)/(0.585-0.516)\approx88\%$ of in-domain gain without CounterFact training labels, supporting cross-benchmark reuse of routing supervision.

With per-item external and parametric costs $c_e,c_p$, write cost is $C(b)=Nc_e+\lfloor bN\rfloor c_p$. Adapter training dominates $c_e$, so the 20\% operating point cuts the dominant term fivefold; combined with the equal-accuracy comparisons above, the ordering converts budget into savings at matched quality.

\subsection{Scale and sequential allocation}
In the all-item oracle evaluation, the fraction of CounterFact updates benefiting from both over retrieval rises from 19.6\% at 0.6B to 47.2\% at 4B. The profitable write set therefore depends on model scale: a fixed write fraction cannot express the same allocation trade-off for both models. Within this 500-item pool, a 20\% budget covers at most $100/236\approx42\%$ of beneficial updates. Even perfect benefit detection must therefore be followed by prioritization among useful writes.

Over 30 CounterFact/4B sequential replays (three seeds, ten orderings), adaptive budget pacing reaches 0.75764 at 20\% budget versus 0.75800 for full-pool selection, a 0.036 pp gap, indicating that per-item priorities transfer to sequential arrival orders.

The selective-write regime is also stable across configurations: on 0.6B the beneficial fraction stays within 17--21\% across LoRA ranks $\{4,8,16,32\}$ and packing densities $\{4,8,16,32\}$ facts/adapter, and adapter training is non-destructive, with only 1\% of 300 unrelated probes degraded. Supervision is inexpensive to collect---100 labeled items ($\approx$3.5 GPU-hours) already recover 55--61\% of oracle gain---so one router trains on a small labeled subset and then serves every budget through the shared ranking.

\section{Conclusion}
\label{sec:conclusion}
\allot frames knowledge updates in LLMs as a budgeted hybrid-memory allocation problem: retain facts externally and budget parametric writes using a reusable memory-aware ranking. It recovers 78.4\% of budget-matched oracle gain at 20\% budget on CounterFact/Qwen3-4B, while reducing parametric writes by 80\%. Feature interactions and shared-store transfer results further support selective rather than indiscriminate adaptation. One ranking supports changing budgets without router retraining or threshold recalibration; the same scores also order writes over time. Future work will explore longer LLM update streams and additional model families.

\clearpage
\section{Compliance with Ethical Standards}
This work did not involve human participants or animals, and no ethical approval was required. No new data were collected; all experiments use publicly available language models and factual-editing benchmarks released for research use.

\section{Acknowledgment}
This work was conducted at Baidu Inc. ChatGPT was used for language editing of the manuscript and limited assistance in debugging experimental code. All AI-assisted edits and code changes were reviewed and validated by the authors, who take full responsibility for the content of this work. The authors declare no relevant financial or nonfinancial interests.

\bibliographystyle{IEEEbib}
\bibliography{references}
\end{document}